# New Techniques for Algorithm Portfolio Design

**Matthew Streeter** and **Stephen F. Smith**
The Robotics Institute
Carnegie Mellon University
Pittsburgh, PA 15213
{matts, sfs}@cs.cmu.edu

## Abstract

We present and evaluate new techniques for designing algorithm portfolios. In our view, the problem has both a scheduling aspect and a machine learning aspect. Prior work has largely addressed one of the two aspects in isolation. Building on recent work on the scheduling aspect of the problem, we present a technique that addresses both aspects simultaneously and has attractive theoretical guarantees. Experimentally, we show that this technique can be used to improve the performance of state-of-the-art algorithms for Boolean satisfiability, zero-one integer programming, and A.I. planning.

## 1 Introduction

Many computational problems that arise in the world are NP-hard, and thus likely to be intractable from a worst-case point of view. However, the particular instances of these problems that are actually encountered can often be solved effectively using heuristics that do not have good worst-case guarantees. Typically there are a number of heuristics available for solving any particular NP-hard problem, and there is no one heuristic that performs best on all problem instances. Thus, when solving a particular instance of an NP-hard problem, it is not clear *a priori* how to best make use of the available CPU time.

Specifically, suppose you wish to solve an instance $x$ of a computational problem, and there are $k$ heuristics available for solving it. Each heuristic, when run on instance $x$, will either solve the instance in finite time (e.g., by returning a provably correct "yes" or "no" answer to a decision problem, returning a provably optimal solution to an optimization problem), or will run forever without solving it. When solving $x$, you will in general have some prior knowledge of how each of the $k$ heuristics behaves on other instances of the same computational problem. Naturally, you would like to solve $x$ as quickly as possible.

In this situation, a natural approach would be to label each previously-encountered problem instance with a set of features, and then to use some machine learning algorithm to predict which of the $k$ heuristics will return an answer in the shortest amount of time. However, if we then run the predicted fastest heuristic and it does not yield an answer after some sufficiently large amount of time, we might suspect that the machine learning algorithm's prediction was a mistake, and might try running a different heuristic instead. Alternatively, if the heuristic is randomized, we might try restarting it and running with a fresh random seed.

We refer to the general problem of determining how to solve a problem instance in this setting as *algorithm portfolio design* [5, 6]. As just illustrated, the problem has both a machine learning aspect (predicting which heuristic will solve the instance first) and a scheduling aspect (determining how long to run a heuristic before giving up and trying a different heuristic). Previous work (e.g., [6, 8, 10]) has largely addressed one of the two aspects in isolation (we discuss previous work in detail in §6). In this work, we present an approach that addresses both aspects of the problem simultaneously and has attractive theoretical guarantees.

We note up front that our work does not address all possible aspects of the algorithm portfolio design problem. For example, we ignore the possibility of making scheduling decisions dynamically based on the observed behavior of the heuristics (e.g., if a heuristic has a progress bar that indicates how close it is to solving the instance). We also ignore the possibility of sharing information (e.g., upper and lower bounds on the optimal value of the objective function) between heuristics as they are executing.

## 1.1 Formal setup

We are given as input a set $\mathcal{H}$ of heuristics (i.e., algorithms with potentially large running time) for solving some computational problem. Heuristic $h$, when run on problem instance $x$, runs for $T(h, x)$ time units before solving the problem. If $h$ is randomized, then $T(h, x)$ is a random variable whose outcome depends on the sequence of random bits supplied as input to $h$.

We will be interested in interleaving the execution of heuristics according to *schedules* of the following form.

**Definition (schedule).** *A schedule $S = \langle (h_1, \tau_1), (h_2, \tau_2), \ldots \rangle$ is a sequence of pairs $(h, \tau) \in \mathcal{H} \times \mathbb{R}_{>0}$, where each pair $(h, \tau)$ represents running heuristic $h$ for time $t$.*

When interpreting a schedule, we allow each heuristic $h \in \mathcal{H}$ to be executed in one of two models (the choice of model need not be the same for all heuristics). If $h$ is executed in the *suspend-and-resume model*, then a pair $(h, \tau)$ represents continuing a run of heuristic $h$ for an additional $\tau$ time units. The run of $h$ is then temporarily suspended and kept resident in memory, to be potentially resumed later on. In contrast, if $h$ is executed in the restart model, then a pair $(h, \tau)$ represents running $h$ from scratch for time $\tau$, and then deleting the run from memory (if $h$ is randomized, the run is performed with a fresh random seed).

Abusing notation slightly, we use $T(S, x)$ to denote the time required to solve problem instance $x$ using schedule $S$. We illustrate the definition of $T(S, x)$ with an example. Consider the schedule

$$S = \langle (h_1, 2), (h_2, 2), (h_1, 4), \ldots \rangle$$

illustrated in Figure 1. Suppose $\mathcal{H} = \{h_1, h_2\}$, both heuristics are deterministic, and $T(h_1, x) = T(h_2, x) = 3$. Then $T(S, x) = 5$ if $h_1$ is executed in the suspend-and-resume model, whereas $T(S, x) = 7$ if $h_1$ is executed in the restart model. Note that in calculating $T(S, x)$ when $S$ is executed in the suspend-and-resume model, we ignore any overhead associated with context-switching.

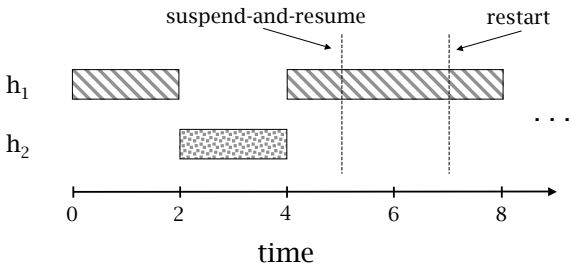

Figure 1: Value of $T(S, x)$ in two execution models.

This class of schedules is quite flexible, and includes *restart-schedules* [11] and *task-switching schedules* [13] as special cases. A restart schedule is a schedule for a single randomized heuristic, executed in the restart model. A task-switching schedule is a schedule for a set of one or more deterministic heuristics, each executed in the suspend-and-resume model.

An *algorithm portfolio* is a way to decide what schedule to use to solve a particular problem instance.

**Definition (algorithm portfolio).** *An algorithm portfolio is a procedure $\phi$ that, given a problem instance $x$, returns a schedule $\phi(x)$ to use to solve $x$.*

We measure the performance of a schedule $S$ on a problem instance $x$ in terms of $\mathbb{E}[T(S, x)]$, where the expectation is over the random bits used in the runs that $S$ performs. We are interested in optimizing this objective in two settings: offline and online.

In the offline setting, we are given as input a set of training instances, along with the value of $T(x, h)$ (or in general, an estimate of its distribution) for each heuristic $h$ and training instance $x$. Our goal is to construct an algorithm portfolio (within some class) that performs optimally on the set of training instances. We would then use such a portfolio to solve additional, similar problem instances more efficiently.

In the online setting, we are fed a sequence $\mathcal{X} = \langle x_1, x_2, \ldots, x_n \rangle$ of problem instances one at a time and must obtain a solution to each instance (via some schedule) before moving on to the next instance. When selecting a schedule $S_i$ to use to solve instance $x_i$, we have knowledge of the previous instances $x_1, x_2, \ldots, x_{i-1}$ but we have no knowledge of $x_i$ itself or of any subsequent instances. In this setting, our goal is to learn an effective algorithm portfolio on-the-fly, again with the aim of minimizing average CPU time.

## 1.2 Summary of results

In §2, we review recent results on a pure scheduling approach to the algorithm portfolio design problem. For the offline setting, the main result is a greedy algorithm that returns a 4-approximation to the optimal schedule; achieving a $4 - \epsilon$ approximation for any $\epsilon > 0$ is NP-hard. For the online setting, the main result is an online schedule-selection algorithm whose worst-case performance guarantees converge to those of the offline greedy approximation algorithm, asymptotically as the number of instances grows large. Note that the latter guarantee does not require any statistical assumptions about the sequence of problem instances.

In §3, we discuss how the online algorithm discussed in §2 can be combined with algorithms for solving the so-called *sleeping experts problem* in order to take advantage of Boolean features of an instance when selecting a schedule. This approach yields an online algorithm that, simultaneously for each feature $f$, is guaranteed to perform near-optimally (i.e., average CPU time asymptotically at most 4 times that of any schedule) on the subset of instances for which $f$ is true.

In §4, we evaluate these techniques experimentally, and show that they can be used to improve the performance of state-of-the-art heuristics for Boolean satisfiability, A.I. planning, and zero-one integer programming.

The results just described apply only to the objective of minimizing average CPU time. In §5, we consider the case in which each heuristic is an *anytime algorithm* that returns solutions of increasing quality over time. We describe how our results for minimizing average CPU time can be generalized to yield schedules with good anytime behavior, and demonstrate the power of this approach by applying it to state-of-the-art heuristics for zero-one integer programming.

## 2 Background

In this section we review recent results on a pure scheduling approach to algorithm portfolio design. These results form the basis of the algorithms and experimental results presented in the rest of the paper.

### 2.1 Offline greedy approximation algorithm

Suppose we collect a set of training instances $\mathcal{X}$, and wish to compute the schedule that performs optimally over the training instances (i.e., the schedule $S$ that minimizes $\sum_{x \in \mathcal{X}} \mathbb{E}\left[T\left(S, x\right)\right]$). We assume that for each heuristic $h \in \mathcal{H}$ and training instance $x \in \mathcal{X}$, the distribution of $T(h, x)$ is known exactly (in practice, we would have to estimate it by performing a finite number of runs).

Building on previous work on the MIN-SUM SET COVER problem [3], Streeter *et al.* [16, 17] developed a greedy approximation algorithm for this offline problem. Let $f(S)$ denote the sum, over all instances $x \in \mathcal{X}$, of the probability that executing schedule $S$ yields a solution to instance $x$. The schedule $G = \langle g_1, g_2, \ldots \rangle$ returned by the greedy approximation algorithm can be defined inductively as follows: $G_1 = \langle \rangle$, $G_j = \langle g_1, g_2, \ldots, g_{j-1} \rangle$ for $j > 1$, and

$$g_j = \underset{a=(h,\tau) \in \mathcal{H} \times \mathbb{R}_{>0}}{\arg\max} \left\{ \frac{f(G_j + a) - f(G_j)}{\tau} \right\} \quad (1)$$

where $G_j + a$ denotes the schedule obtained by appending the pair $a$ to $G_j$.[1] Informally, $G$ is constructed by greedily appending a run $a = (h, \tau)$ to the schedule so as to maximize the expected number of instances $a$ solves per unit time.

---
[1] Evaluating the arg max in (1) requires considering

The performance of $G$ is summarized by the following theorem. The theorem shows that, assuming P $\neq$ NP, the greedy schedule has optimal worst-case performance from an approximation standpoint (among schedules that can be computed in polynomial time).

**Theorem 1** (Streeter *et al.*, 2007a; 2007b). *$G$ is a 4-approximation to the optimal schedule. That is,*

$$\sum_{x \in \mathcal{X}} \mathbb{E}\left[T\left(G, x\right)\right] \leq 4 \cdot \min_S \left\{ \sum_{x \in \mathcal{X}} \mathbb{E}\left[T\left(S, x\right)\right] \right\} \ .$$

*Furthermore, for any $\epsilon > 0$, obtaining a $4 - \epsilon$ approximation to the optimal schedule is NP-hard (even in the special case where all heuristics are deterministic).*

### 2.2 Online greedy algorithm

In the online setting, a sequence $\langle x_1, x_2, \ldots, x_n \rangle$ of problem instances arrive one at a time, and one must solve each instance $x_i$ via some schedule (call it $S_i$) before moving on to instance $x_{i+1}$. When selecting $S_i$, one has no knowledge of $x_i$ itself. After solving $x_i$, one learns only the outcomes of the runs that were actually performed when executing $S_i$. As in the offline setting, the goal is to minimize the average CPU time required to solve each instance in the sequence.

Recently, Streeter and Golovin [15] developed an online algorithm for an abstract scheduling problem that includes this online problem as a special case. For the results of [15] to apply, we must make some additional assumptions. First, we assume that $T(h, x_i)$ is an integer for all heuristics $h$ and instances $x_i$. Second, we assume that the CPU time the online algorithm uses up on any particular instance $x_i$ is artificially capped at some value $B$ (without such a cap, the online algorithm could be forced to spend an arbitrarily large amount of CPU time solving a single instance, and we could prove no meaningful bounds on its performance).

The algorithm presented in [15] is called **OG**, for "online greedy", and can be viewed as an online version of the greedy approximation algorithm described in §2.1. The following theorem shows that its worst-case performance guarantees approach those of the offline greedy algorithm, asymptotically as the number of problem instances approaches infinity. The theorem can be proved as a corollary of [15, Theorem 11] (for a formal derivation, see [14, Chapter 3]).

---
$O(r |\mathcal{X}|)$ values of $\tau$ per heuristic, where $r$ is the maximum number of runs used to estimate the distribution of $T(h, x)$. For more details, see [14].

**Theorem 2** (Streeter and Golovin, 2007). *Algorithm* **OG** *[15], run with exploration probability* $\gamma = \Theta\left(n^{-\frac{1}{4}}\right)$, *has the following guarantee. Let* $T_i = \min\{B, T(S_i, x_i)\}$, *for some* $B > 0$. *Then*

$$\sum_{i=1}^{n} \mathbb{E}[T_i] \leq 4 \cdot \min_{S \in \mathcal{S}} \left\{ \sum_{i=1}^{n} \mathbb{E}[T(S, x)] \right\} + O\left(n^{\frac{3}{4}}\right) .$$

## 3 Exploiting Features

The algorithms referred to in theorems 1 and 2 provide no mechanism for tailoring the choice of schedule to the particular problem instance being solved. In practice, there may be quickly-computable features that distinguish one instance from another and suggest the use of different heuristics. In this section, we describe how existing techniques for solving the so-called *sleeping experts problem* can be used to exploit such features in an attractive way.

The sleeping experts problem is defined as follows. One has access to a set of $M$ experts. On each day, a given expert is either *awake*, in which case the expert dispenses a piece of advice, or the expert is *asleep*. At the beginning of day $i$, one must select an awake expert whose advice to follow. Following the advice of expert $j$ on day $i$ incurs a loss $\ell_j^i \in [0, 1]$. At the end of day $i$, the value of the loss $\ell_j^i$ for each (awake) expert $j$ is made public, and can be used as the basis for making choices on subsequent days. Note that the historical performance of an expert does not imply any guarantees about its future performance. Remarkably, randomized expert-selection algorithms nevertheless exist that achieve the following guarantee: simultaneously for each $j$, one's expected loss on the subset $D_j$ of days when $j$ was awake is at most $\sum_{i \in D_j} \ell_j^i + O\left(\sqrt{n \log M} + \log M\right)$. Thus, when using such an algorithm[2], one asymptotically performs as well as any fixed expert on the subset of days that expert was awake.

Suppose that each problem instance $x_i$ is labeled with the values of $M$ Boolean features. We will exploit such features by applying the sleeping experts algorithm in a standard way. We create, for each feature $j$, a copy $\mathcal{A}_j$ of the online schedule-selection algorithm **OG** that is only run on instances where feature $j$ is true. We then use an algorithm for the sleeping experts problem to select among the schedules returned by the various copies, as described in the pseudo-code for **OG**$^{se}$. Due to space constraints, the pseudo-code refers to [15, 17]

[2]See [2] for a description of such an algorithm. The algorithm maintains, for each expert, a weight that is adjusted based on its performance relative to other experts. On each day, experts are selected with probability proportional to their weights.

for the details of certain steps. As in §2.2, we use $B$ to denote an artificial bound on CPU time.

---

**Algorithm OG$^{se}$**

Initialization: let $\mathcal{E}$ be a copy of the sleeping experts algorithm of [2]; and for each feature $j$, let $\mathcal{A}_j$ be a copy of **OG** [15].

For $i$ from 1 to $n$:
1. Let $F_i$ be the set of features that are true for $x_i$. For each feature $j \in F_i$, use $\mathcal{A}_j$ to select a schedule $S_{i,j}$.

2. Use $\mathcal{E}$ to select a feature (expert) $j_i \in F_i$, and select the schedule $S_i = S_{i,j_i}$.

3. With probability $\gamma = \Theta\left(n^{-\frac{1}{4}}\right)$, *explore* as follows. Using the procedure of [17], run each heuristic for time $O(B \log B)$ in order to obtain a function $\hat{f}$ such that for any schedule $S$, $\mathbb{E}\left[\hat{f}(S)\right] = \mathbb{E}[\min\{B, T(S, x_i)\}]$. Feed $\hat{f}$ back to each $\mathcal{A}_j$, as described in [15]. Finally, for each $j$, set $\ell_j^i = \frac{1}{B}\hat{f}(S_{i,j})$. Otherwise (with probability $1 - \gamma$) set $\ell_j^i = 0$ for all $j$.

4. For each $j \in F_i$, feed back $\ell_j^i$ to $\mathcal{E}$ as the loss for expert $j$.

---

The performance of **OG**$^{se}$ is summarized by the following theorem.

**Theorem 3.** *Let* $\mathcal{X}_j$ *be the subset of instances for which feature $j$ is true. Let* $T(x)$ *be the CPU time spent by* **OG**$^{se}$ *on instance $x$. Then, simultaneously for each $j$, we have*

$$\mathbb{E}\left[\sum_{x \in \mathcal{X}_j} T(x)\right] \leq 4 \cdot \min_{S} \left\{\sum_{x \in \mathcal{X}_j} \mathbb{E}[T(S, x)]\right\} + O\left(n^{\frac{3}{4}}\right) .$$

*Proof.* As already discussed, the algorithm $\mathcal{E}$ used as a subroutine in **OG**$^{se}$ guarantees that, for any $j$,

$$\sum_{x \in \mathcal{X}_j} \ell_{j_i}^i \leq \sum_{x \in \mathcal{X}_j} \ell_j^i + R \quad (2)$$

where $R = O\left(\sqrt{n \log M} + \log M\right)$. Define $L_i(S) = \mathbb{E}[\min\{B, T(S, x_i)\}]$. Thus $\mathbb{E}\left[\ell_j^i\right] = \frac{\gamma}{B} L_i(S_{i,j})$. Taking the expectation of both sides of (2) yields

$$\sum_{x \in \mathcal{X}_j} L_i(S_i) \leq \sum_{x \in \mathcal{X}_j} L_i(S_{i,j}) + \frac{B}{\gamma} R .$$

Note that $\frac{B}{\gamma}R = O\left(n^{\frac{3}{4}}\right)$ (for constant $M$). At the same time, by Theorem 2 we have

$$\sum_{x \in \mathcal{X}_j} L_i(S_{i,j}) \le 4 \cdot \min_S \left\{ \sum_{x \in \mathcal{X}_j} \mathbb{E}\left[T(S, x)\right] \right\} + O\left(n^{\frac{3}{4}}\right).$$

Finally, because $\gamma = \Theta\left(n^{-\frac{1}{4}}\right)$, we have $\mathbb{E}\left[\sum_{x \in \mathcal{X}_j} T(x)\right] \le \sum_{x \in \mathcal{X}_j} L_i(S_i) + O\left(n^{\frac{3}{4}}\right)$. Putting these equations together proves the theorem. □

Note that Theorem 3 provides a very strong guarantee. For example, if each instance is labeled as either "large" or "small" and also as either "random" or "structured", then the performance of $\mathbf{OG}^{se}$ on large instances will be nearly as good as that of the optimal schedule for large instances, and *simultaneously* its performance on structured instances will be nearly as good as that of the optimal schedule for structured instances (even though these subsets of instances overlap, and the optimal schedule for each subset may be quite different).

## 4  Experimental Evaluation

In this section, we evaluate the algorithms presented in the previous section experimentally using data from recent solver competitions.

### 4.1  Solver competitions

Each year, various computer science conferences hold competitions designed to assess the state of the art solvers in some problem domain. In these competitions, each submitted solver is run on a sequence of problem instances, subject to some per-instance time limit. Solvers are awarded points based on the instances they solve and how fast they solve them, and prizes are awarded to the highest-scoring solvers.

The experiments reported here make use of data from the following three solver competitions.

1. **SAT 2007**. Boolean satisfiability is the task of determining whether there exists an assignment of truth values to a set of Boolean variables that satisfies each clause (disjunction) in set of clauses. SAT solvers are used as subroutines in state-of-the-art algorithms for hardware and software verification and A.I. planning. The SAT 2007 competition included industrial, random, and hand-crafted benchmarks.

2. **IPC-5**. A.I. planning is the problem of finding a sequence of actions (called a plan) that leads from a starting state to a desired goal state, according to some formal model of how actions affect the state of the world. We used data from the *optimal planning* track of the Fifth International Planning Competition (IPC-5), in which the model of the world is specified in the STRIPS language and the goal is to find a plan with (provably) minimum length.

3. **PB'07.** Pseudo-Boolean optimization is the task of minimizing a function of zero-one variables subject to algebraic constraints, also known as zero-one integer programming. On many benchmarks, pseudo-Boolean optimizers (which are usually based on SAT solvers) outperform general integer programming packages such as CPLEX [1]. The PB'07 evaluation included both optimization and decision (feasibility) problems from a large number of domains, including formal verification and logic synthesis.

Our experiments for each solver competition followed a common procedure. First, we determined the value of $T(h, x)$ for each heuristic $h$ and benchmark instance $x$ using data available on the competition web site (we did not actually run any of the heuristics). The heuristics considered in these competitions are deterministic (or randomized, but run with a fixed random seed), so $T(h, x)$ is simply a single numeric value. If a heuristic did not finish within the competition time limit, then $T(h, x)$ is undefined. Second, we discarded any instances that none of the heuristics could solve within the time limit.

Given a schedule $S$ and instance $x$, we will not generally be able to determine the true value of $T(S, x)$, due to the fact that $T(h, x)$ is undefined for some heuristics. We can, however, determine the value of $\min\{B, T(S, x)\}$, where $B$ is the competition time limit. We use this lower bound in all the comparisons that follow.

### 4.2  Number of training instances required in practice

In this section we investigate how the number of available training instances affects the quality of a schedule computed using those training instances. To do so, we adopted the following procedure. Given a set of $n$ instances, we select $m < n$ training instances at random, then use the greedy algorithm from §2.1 to compute an approximately optimal schedule[3] for the training instances. We then use this schedule to solve each of the $n-m$ remaining instances, and record the average CPU time it requires. We examined all values of $m$ that were powers of 2 less than $n$. For each value of $m$, we repeated the experiment 100 times and averaged the results.

---
[3] For all solver competitions, the number of heuristics was large enough that computing an optimal schedule via dynamic programming was impractical.

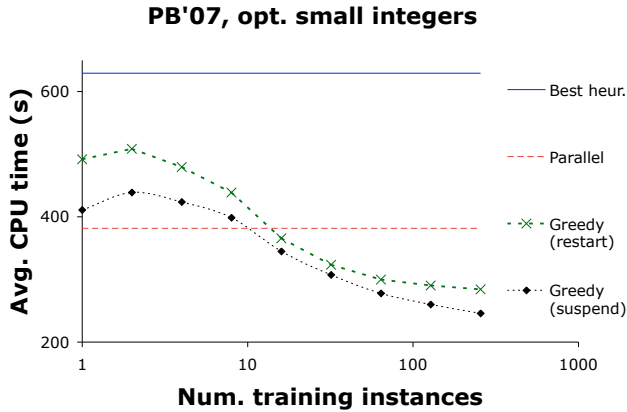

Figure 2: Experimental results for PB'07 data.

Figure 2 depicts the results for optimization problems from the "small integers" track of the PB'07 competition. The figure shows average CPU time (on test instances) as a function of the number of training instances, for both versions of the greedy algorithm (suspend-and-resume and restart). For comparison, the figure also shows the average CPU time required by the fastest individual solver, as well as a schedule that simply ran all solvers in parallel (i.e., if there are $k$ solvers, each one receives a $\frac{1}{k}$ fraction of the CPU time).

Figure 2 has several noteworthy features. First, only a small number of training instances (in this case 16) are required in order to produce a schedule that outperforms both the fastest individual solver and the naïve parallel schedule. Second, with a sufficient number of training instances, the gap between the performance of the greedy schedules and that of the fastest individual solver is significant (in this case, more than a factor of 2). Third, the suspend-and-resume model offers only a relatively small advantage over the restart model. We have observed these same three trends in a number of other cases (e.g., see Figure 3).

We note that previous work (e.g., [16]) gave learning-theoretic bounds on the number of training instances required to learn a near-optimal schedule; however, these worst-case upper bounds are quite pessimistic relative to our experimental results.

### 4.3 Exploiting features

We now examine the benefit of using Boolean features to help decide which schedule to use for solving a particular problem instance. We present results for two instance sets: the *random* category of the SAT 2007 competition, and the optimal planning track of IPC-5. For the SAT instances, we labeled each instance with Boolean features based on the size of the formula, the ratio of clauses to variables, and the number of literals per clause. For the planning instances, we used features based on the planning domain, the number of goals, the number of objects, and the number of predicates in the initial conditions.

To evaluate the effect of features, we used a procedure similar to the one used in the experiments summarized in Figure 2. Given a data set, we sample $m$ training instances at random, and examine how average performance (on test instances) varies as a function of $m$. For each value of $m$, we again repeated the experiment 100 times and averaged the results. In addition to evaluating the greedy algorithm from §2.1, we now evaluate two other approaches. The first approach, which we refer to as "Greedy w/features", uses the algorithm $\mathbf{OG}^{se}$ from §3 to select (suspend-and-resume) schedules as follows. First, we run $\mathbf{OG}^{se}$ on each of the $m$ training instances, with exploration probability $\gamma = 1$. We then run the algorithm on each of the $n-m$ test instances, with exploration probability $\gamma = 0$ (so the algorithm receives no feedback on test instances). The second approach, which we refer to as "Features only" below, is similar except that it uses the sleeping experts algorithm of [2] to select a *single* heuristic (rather than a schedule), and runs that heuristic until it obtains a solution. Here we focus on performance as a function of the number of training instances, because the number of benchmark instances was typically too small to allow for good performance in the online setting of §2.2.

Figures 3 (A) and (B) present our results for the SAT and planning instances, respectively. Both graphs exhibit two noteworthy features. First, when the number of training instances is relatively small, a pure scheduling approach outperforms a purely feature-based approach; but as the number of training instances increases, the reverse is true. This behavior makes intuitive sense: when the number of training instances is small, committing to a single heuristic based on the training data is a very risky thing to do, and thus a purely feature-based approach can perform very poorly (e.g., worse than the naïve parallel schedule); as the number of training instances increases this becomes less of a risk. Second, in all cases, an approach that uses features to select schedules outperforms either a pure scheduling or purely feature-based approach.

Figure 4 depicts the (suspend-and-resume) schedule returned by the greedy algorithm when all available

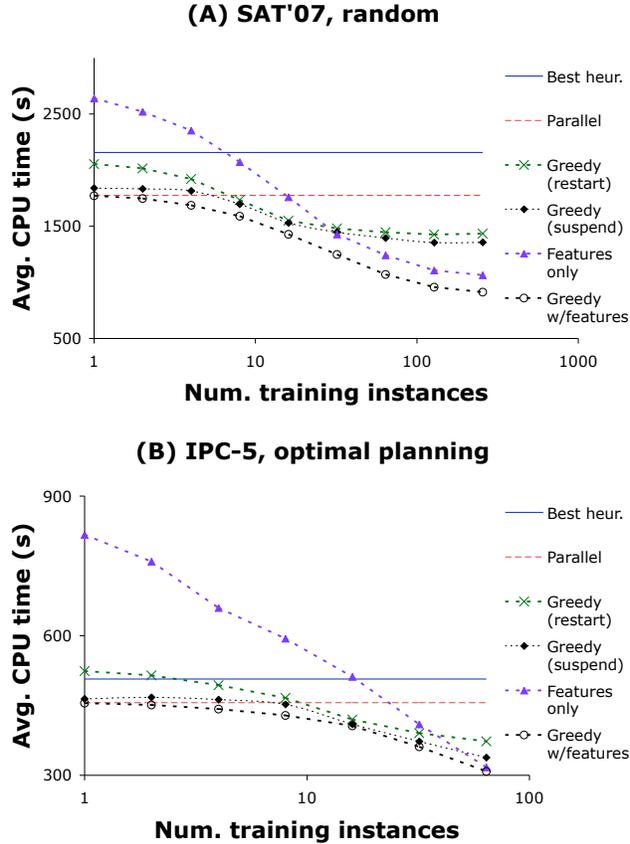

Figure 3: Experimental results for (A) SAT'07 data, *random* category and (B) IPC-5, optimal track.

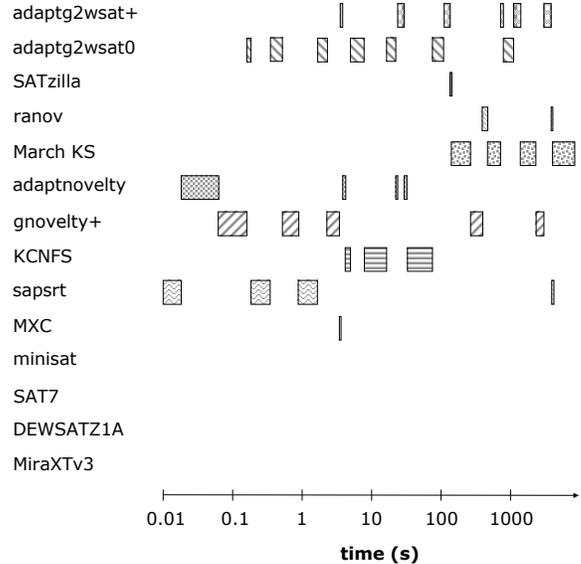

Figure 4: Schedule for SAT'07 data, *random* category. The solvers are listed in ascending order of (the lower bound on) average CPU time.

SAT instances are used as training data. As indicated in the figure, the greedy schedule makes use a variety of different SAT solvers, and spends a significant amount of time running solvers whose overall average CPU time did not put them at the top of the competition.

## 5 Combining Anytime Algorithms

Thus far, we have thought of a heuristic as a program that, given a problem instance, runs for some fixed amount of time before definitively solving it (e.g., by returning a provably optimal solution). Now suppose instead that our heuristics are anytime algorithms that return solutions of increasing quality over time. In this case, we would like to construct a schedule that yields near-optimal solutions quickly, in addition to yielding provably optimal solutions quickly.

One simple way to do this is as follows. Define, for each instance, a set of objectives to achieve (e.g., finding a solution with cost at most $\alpha$ times optimal, for each $\alpha \in \{2, 1.5, 1.01\}$). For simplicity, consider the offline setting described in §2.1. For each training instance $x$, create a new set of fictitious instances $\tilde{x}_1, \tilde{x}_2, \ldots, \tilde{x}_k$, one for each of the $k$ objectives. For each heuristic $h$, define $T(h, \tilde{x}_i)$ to be the time that $h$ requires to achieve the $i^{th}$ objective. Then, the average time a schedule or heuristic takes to "solve" the fictitious instances is simply the average time it takes to achieve each of the $k$ objectives on the original instances. If some objectives are more important than others, we can weight the fictitious instances accordingly (the results described in §2 readily extend to weighted sets of instances).

To evaluate this approach, we revisit the experiments performed in §4 using the PB'07 competition data, but now we measure the performance of a schedule as the average of (i) the time the schedule takes to find a feasible solution, (ii) the time the schedule takes to find an optimal solution, and (iii) the time the schedule takes to prove optimality (or to prove that the problem is infeasible).

Table 1 summarizes the results of these experiments. For each track of the PB'07 competition and for each of the three objectives, we define a *speedup factor* equal to the (lower bound on) average CPU time required by the fastest individual heuristic to achieve that objective, divided by the corresponding quantity for the (suspend-and-resume) greedy schedule, where the greedy algorithm is evaluated under leave-one-out cross-validation. Note that in general, the three different speedup factors listed for each track represent a comparison against three *different* heuristics.

Table 1 shows that for all three tracks, we were able to generate a schedule that *simultaneously* outperformed each of the original heuristics in terms of each of the three objectives we considered. The results of these experiments could potentially be improved by using features[4] as in §4.3, and by sharing upper and lower bounds on the optimal objective function value among heuristics as they are discovered.

Table 1: Speedup factors for experiments with anytime algorithms, using PB'07 data.

| Track | Speedup (prove opt) | Speedup (find opt) | Speedup (find feas) |
|---|---|---|---|
| Sm. ints | 2.5 | 2.9 | 3.7 |
| Sm. ints non-linear | 1.6 | 1.3 | 1.4 |
| Big ints. | 1.2 | 1.5 | 1.4 |

## 6 Related Work

Previous work on algorithm portfolio design has almost always focused on a single aspect of the problem. In particular, almost all previous theoretical work has focused on the scheduling aspect of the problem, whereas the bulk of the experimental work has focused on the machine learning aspect of the problem. We now discuss previous work on each of these two aspects of the problem in greater detail.

### 6.1 Scheduling approaches

A number of papers have considered the problem of coming up with a schedule for allocating time to runs of one or more algorithms.

The earliest work on this problem measured the performance of a schedule in terms of its competitive ratio (i.e., the time required to solve a given problem instance using the schedule, divided by the time required by the optimal schedule for that instance). Results of this work include the universal restart schedule of Luby *et al.* [11] and the schedule of Kao *et al.* [9] for allocating time among multiple deterministic algorithms subject to memory constraints.

Subsequent work focused on developing schedules tailored to a particular class of problems. Gomes *et al.* [7] demonstrated that (then) state-of-the-art heuristics for Boolean satisfiability and constraint satisfaction could be dramatically improved by randomizing the heuristic's decision-making heuristics and running the randomized heuristic with an appropriate restart schedule. Huberman *et al.* [8] and Gomes *et al.* [6] combined multiple algorithms into a portfolio by running each algorithm in parallel at equal strength and assigning each algorithm a fixed restart threshold.

To fully realize the power of this approach, one must solve the problem of computing a schedule that performs well on average over a given set of problem instances collected as training data. Independently, Petrik and Zilberstein [12] and Sayag *et al.* [13] addressed this problem for two classes of schedules: task-switching schedules and resource-sharing schedules. For each of these two classes of schedules, the problem of computing an optimal schedule is NP-hard, and accordingly their algorithms have exponential running time (as a function of the number of algorithms being scheduled). Recently, Streeter *et al.* [16] presented a polynomial-time 4 approximation algorithm for computing task-switching schedules, as reviewed in §2.1.

### 6.2 Machine learning approaches

Another approach to algorithm portfolio design is to use features of instances to attempt to predict which algorithm will run the fastest on a given instance, and then simply run that algorithm exclusively. As an example of this approach, Leyton-Brown *et al.* [10] use least squares regression to estimate the running time of each algorithm based on quickly-computable instance features, and then run the algorithm with the smallest predicted running time. Xu *et al.* [18] presented an improved version of this approach that used a two-step prediction scheme in which the answer to a decision problem is predicted using a binary classifier, and run times are then estimated conditioned on the classifier's prediction.

### 6.3 Integrated approaches

In addition to the work just described, there has been previous work that addresses both the scheduling and machine learning aspects of the algorithm portfolio design problem simultaneously. For example, Gagliolo and Schmidhuber [4] presented an approach for allocating CPU time among heuristics in an online setting, based on statistical models of the behavior of the heuristics. Although their approach has no rigorous performance guarantees and would not perform well in the worst-case online setting considered in this paper, it would be interesting to compare their approach to ours experimentally.

## 7 Conclusions

This paper presented a new technique for addressing the scheduling and machine learning aspects of the algorithm portfolio design problem, and evaluated the technique experimentally. Our main experimental findings can be summarized as follows.

---

[4]We do not present experiments that use features in conjunction with the PB'07 data because we could not readily find a suitable set of features.

1. In a number of well-studied problem domains, existing state-of-the-art heuristics can be combined into a new and faster heuristic simply by collecting a few dozen training instances and using them to compute a schedule for interleaving the execution of the existing heuristics.

2. State-of-the-art anytime algorithms for solving optimization problems can be combined, via a schedule, into an algorithm with better anytime performance.

3. Instance-specific features can be used to generate a custom schedule for a particular problem instance. Using this approach can result in better performance than using either a pure scheduling approach or a purely feature-based approach.

As suggested in §1, our experimental results could potentially be improved in at least two ways. First, we could attempt to predict a heuristic's remaining running time based on its current state and adapt our schedule accordingly. Second, we could share information among heuristics during the process of solving an instance (e.g., when solving optimization problems, the heuristics could share upper and lower bounds on the optimal objective function value).

## Acknowledgements

Many thanks to Avrim Blum for suggesting the use of a sleeping experts algorithm in our problem setting. This research was supported in part by DARPA under Contract # FA8750-05-C-0033, and by the CMU Robotics Institute.

## References

[1] Fadi A. Aloul, Arathi Ramani, Igor L. Markov, and Karem A. Sakallah. Generic ILP versus specialized 0-1 ILP: An update. In *ICCAD*, pages 450–457, 2002.

[2] Avrim Blum and Yishay Mansour. From external to internal regret. *Journal of Machine Learning Research*, 8:1307–1324, 2007.

[3] Uriel Feige, László Lovász, and Prasad Tetali. Approximating min sum set cover. *Algorithmica*, 40(4):219–234, 2004.

[4] Matteo Gagliolo and Jürgen Schmidhuber. Dynamic algorithm portfolios. In *AIMATH*, 2006.

[5] Carla P. Gomes and Bart Selman. Algorithm portfolio design: Theory vs. practice. In *UAI*, pages 190–197, 1997.

[6] Carla P. Gomes and Bart Selman. Algorithm portfolios. *Artificial Intelligence*, 126:43–62, 2001.

[7] Carla P. Gomes, Bart Selman, and Henry Kautz. Boosting combinatorial search through randomization. In *AAAI*, pages 431–437, 1998.

[8] Bernardo A. Huberman, Rajan M. Lukose, and Tad Hogg. An economics approach to hard computational problems. *Science*, 275:51–54, 1997.

[9] Ming-Yang Kao, Yuan Ma, Michael Sipser, and Yiqun Yin. Optimal constructions of hybrid algorithms. In *SODA*, pages 372–381, 1994.

[10] Kevin Leyton-Brown, Eugene Nudelman, Galen Andrew, James McFadden, and Yoav Shoham. Boosting as a metaphor for algorithm design. In *CP*, pages 899–903, 2003.

[11] Michael Luby, Alistair Sinclair, and David Zuckerman. Optimal speedup of Las Vegas algorithms. *Information Processing Letters*, 47:173–180, 1993.

[12] Marek Petrik and Shlomo Zilberstein. Learning parallel portfolios of algorithms. *Annals of Mathematics and Artificial Intelligence*, 48(1-2):85–106, 2006.

[13] Tzur Sayag, Shai Fine, and Yishay Mansour. Combining multiple heuristics. In *STACS*, pages 242–253, 2006.

[14] Matthew Streeter. *Using Online Algorithms to Solve NP-Hard Problems More Efficiently in Practice*. PhD thesis, Carnegie Mellon University, 2007.

[15] Matthew Streeter and Daniel Golovin. An online algorithm for maximizing submodular functions. Technical Report CMU-CS-07-171, Carnegie Mellon University, 2007.

[16] Matthew Streeter, Daniel Golovin, and Stephen F. Smith. Combining multiple heuristics online. In *AAAI*, pages 1197–1203, 2007.

[17] Matthew Streeter, Daniel Golovin, and Stephen F. Smith. Restart schedules for ensembles of problem instances. In *AAAI*, pages 1204–1210, 2007.

[18] Lin Xu, Frank Hutter, Holger H. Hoos, and Kevin Leyton-Brown. SATzilla07: The design and analysis of an algorithm portfolio for SAT. In *CP*, pages 712–727, 2007.